\documentclass[sensors,article,accept,moreauthors,pdftex]{Definitions/mdpi}
\firstpage{1} 
\pubvolume{1}
\issuenum{1}
\articlenumber{0}
\pubyear{2024}
\copyrightyear{2024}
\externaleditor{Academic Editor: Name}
\datereceived{8 June 2024} 
\daterevised{28 June 2024} 
\dateaccepted{ } 
\datepublished{ } 
\hreflink{https://doi.org/} 

\usepackage[utf8]{inputenc}
\usepackage{mathtools}
\usepackage{algorithmic}
\usepackage[ruled]{algorithm2e}
\usepackage{amsmath,amssymb,amsfonts}
\usepackage{algorithmic}
\usepackage{graphicx}
\usepackage{epstopdf}
\usepackage{color}
\renewcommand{\hl}[1]{#1}
\Title{Joint Optimization of Age of Information and Energy Consumption in NR-V2X System Based on Deep \linebreak Reinforcement Learning}

\TitleCitation{Joint Optimization of Age of Information and Energy Consumption in NR-V2X System Based on Deep Reinforcement Learning}

\Author{\hl{Shulin Song} 
 $^{1,2,\dagger}$, Zheng Zhang $^{1,2,\dagger}$, Qiong Wu $^{1,2,}$*, Pingyi Fan $^{3}$ and Qiang Fan $^{4}$ }

 \AuthorNames{Shulin Song, Zheng Zhang, Qiong Wu, Pingyi Fan   and Qiang Fan}

\AuthorCitation{Song, S.; Zhang, Z.; Wu, Q.; Fan, P.; Fan, Q.}

\address{$^{1}$ \quad School of Internet of Things Engineering, Jiangnan University, Wuxi 214122, China;  songshulin@jiangnan.edu.cn (S.S.); zhengzhang@stu.jiangnan.edu.cn (Z.Z.)\\
	$^{2}$ \quad Zhuhai Fudan Innovation Institute, Zhuhai 519031, China\\
	$^{3}$ \quad Department of Electronic Engineering, Beijing National Research Center for Information Science and Technology, Tsinghua University, Beijing 100084, China;  \hl{fpy@tsinghua.edu.cn} 
 \\
     $^{4}$ \quad Qualcomm, San Jose, CA 95110, USA;   \hl{qf9898@gmail.com}
}	
\corres{Correspondence: qiongwu@jiangnan.edu.cn; Tel.: +86-0510-8591-0633}
\firstnote{These authors contributed equally to this work.}	
		
\abstract{As autonomous driving may be  the most important application scenario of the next 
generation, the development of wireless access technologies enabling reliable and low-latency vehicle communication becomes crucial. To address this, 3GPP has developed Vehicle-to-Everything (V2X) specifications based on 5G New Radio (NR) technology, where Mode 2 Side-Link (SL) communication resembles Mode 4 in LTE-V2X, allowing direct communication between vehicles. This supplements SL communication in LTE-V2X and represents the latest advancements in cellular V2X (C-V2X) with the improved performance of NR-V2X. However, in NR-V2X Mode 2, resource collisions still occur and thus degrade the 
age of information (AOI). Therefore, 
an 
interference cancellation method is employed to mitigate this impact 
{ by combining NR-V2X with Non-Orthogonal multiple access (NOMA) technology}. In NR-V2X, when vehicles select smaller resource reservation intervals (RRIs), higher-frequency transmissions 
{use} more energy to reduce AoI. Hence, it is 
{important to} jointly consider
AoI and communication energy consumption based on NR-V2X communication. Then, we formulate such an optimization problem and employ the Deep Reinforcement Learning (DRL) algorithm  to compute the optimal transmission RRI and transmission power for each transmitting vehicle to reduce the energy consumption of each transmitting vehicle and the AoI of each receiving vehicle. 
{Extensive simulations demonstrate the performance of our proposed algorithm.}
}

\begin{document}
		
		\section{Introduction}
		\label{sec1}
		As autonomous driving is one of the most promising application fields for the next generation {of} communication systems, the~development of
		reliable and low-latency vehicle communication becomes crucial. Such technologies not only enhance interconnectivity between vehicles but also facilitate efficient communication between vehicles and infrastructure. 
		{For} autonomous driving vehicles, wireless access technologies play a critical role by providing real-time information exchange and collaboration capabilities among vehicles, thereby enhancing driving safety and efficiency.
		Therefore, continuous innovation and development in wireless access technologies hold strategic significance, further propelling the advancement and adoption of autonomous driving vehicle technology~\cite{wu2024Fog,wu2016performance}.
		In recent related research, the~scheduling and allocation of factors that affect communication performance are the main research directions~\cite{chen2008unified,zhang2004adaptive,wu2024urllc,wu2023high,jing2010optimal}. The~main research method focuses on DRL~\cite{qiong2023towards,wu2022mobility,wu2024cooperative}.
		
		3GPP has formulated V2X specifications based on 5G NR technology to support ultra-low latency and ultra-high reliability in evolving vehicle applications, communications, and~service requirements~\cite{Tlake2021}. 
		As pointed out in~\cite{Garcia2021}, the~development of SL in NR-V2X is to supplement and expand the SL communication in LTE-V2X. However, the~autonomous resource {allocation} 
		method used in Mode 2 still suffers from resource collisions
		{. When} the RRI decreases and the vehicle occupies more resources, the~
		{collision probability}
		gradually increases. {Collisions mean one vehicle receiving multiple messages simultaneously, which results in the~mutual interference between these messages and~directly reduces the Signal-to-Interference-plus-Noise Ratio (SINR) of each message,  which degrades the transmission time. }
		Therefore, NOMA is used to mitigate the impact of this situation. 
		{When the vehicle receives multiple collision messages, NOMA decodes these messages separately to increase the SINR of messages with relatively low power and improve communication performance.}
		
		Furthermore, as~mentioned in~\cite{Hu2018}, while ensuring communication effectiveness, energy consumption also needs to be considered. The~transmission power also affects the SINR and energy consumption during the transmission process~\cite{Wijerathna2021}. When the power is high, the~SINR is more likely to meet the requirements for successful transmission, thereby reducing AoI, but~energy consumption will also increase 
		at the same time~\cite{zhang2023efficient}. Therefore, in~NR-V2X communication, it is necessary to comprehensively consider the balance between communication effectiveness and energy consumption.	
		In order to address this issue 
		, a~resource allocation scheme based on DRL is proposed to allocate RRI and power for vehicles
		ensuring low energy consumption and low information age of the system during the communication \hl{process} 
 (The source code has been released at the following link: {\url{https://github.com/qiongwu86/Joint-Optimization-of-AoI-and-Energy-Consumption-in-NR-V2X-System-based-on-DRL}} \hl{accessed on}
). 
		The performance of our proposed resource allocation method is evaluated through simulation experiments, and~the results demonstrate that it can improve the communication performance of the NR-V2X vehicle networking~system.		
		
		The remainder of this paper is structured as follows: Section \ref{sec2} provides a review of the related literature. In~Section \ref{sec3}, we introduce the system model and formulate the optimization problem. Section \ref{sec4} simplifies the formulated optimization problem and presents a near-optimal solution using DRL. We conduct simulations to demonstrate the effectiveness of our proposed method in Section \ref{sec5}, followed by concluding remarks\linebreak  in Section~\ref{sec6}.
		
		\section{Related~Work}\label{sec2}
		In this section, we {have reviewed} 
		the existing work on the analysis
		, network optimization, and the application of reinforcement learning in NR-V2X systems.
		Rehman~et~al. proposed an analytical model for evaluating the NR-V2X communication performance, focusing on the sensor-based semi-persistent scheduling operations defined in NR-V2X Mode 2 and comparing them with LTE-V2X Mode 4. {For} 
		different physical layer specifications, the~average packet success probability for LTE-V2X and NR-V2X was analyzed, and a moment matching approximation method was used to approximate the SINR statistics under the Nakagami-lognormal composite channel model. It was shown that, under conditions of relatively large inter-vehicle spacing and a high number of vehicles, NR-V2X outperforms LTE-V2X~\cite{rehman2023}.
		Anwar~et~al. evaluated and compared the PHY layer performance of various V2V communication technologies. The~results showed that NR-V2V outperforms other standards (such as IEEE 802.11bd) in terms of reliability, range, delay, and~data rate. However, under~the same modulation and coding scheme, IEEE 802.11bd performs better in terms of PER. Although~the lowest MCS of NR-V2X is more reliable than IEEE 802.11bd, IEEE 802.11bd has a wider range. Overall, NR-V2X performs best in V2V communication~\cite{anwar2019}.
		
	Ref.	{\cite{gong2018} investigated the energy consumption and AoI of a device that is a single-source node,}
		considering potential transmission failures due to poor channel conditions from the source node to the receiver. For~a threshold-based retransmission strategy in the system, the~corresponding closed-form expressions for average AoI and energy consumption were derived, which can be used  to estimate channel failure probabilities and maximum retransmission attempts. 
		Authors of~\cite{gu2019} adopted the Truncated Automatic Repeat Request  scheme, where terminal devices repeatedly send the current status update until reaching the maximum allowable transmission attempts or generating a new status update. Closed-form expressions for average AoI, average peak AoI, and~average energy consumption are derived based on the evolution process of AoI.
		Authors of~\cite{yin2020} primarily considered scenarios where multiple information sources are needed to transmit information for completing status updates, thereby reducing AoI. It investigated the problem of packet scheduling based on information freshness for application-oriented scenarios with correlated multiple information sources. Specifically, it {employs} 
		AoI to characterize the freshness of status updates for applications and {formulates} 
		the application-oriented scheduling problem as an MDP problem, utilizing DRL for~solving.
		
		Liang~et~al. proposed an implementation method for an Integrated Sensing and Communication (ISAC) system for vehicular networks, addressing the potential performance degradation caused by the coexistence of millimeter-wave radar and communication by extending NR-V2X Mode 2. By~using semi-persistent scheduling (SPS) resource selection and dynamically adjusting the radar scanning cycle and transmission power of each vehicle based on the speed and channel congestion status reported by neighboring vehicles, the~ISAC system ensures that high-priority vehicles occupy spectrum resources preferentially. Simulation results validated the effectiveness of this approach in improving radar and communication performance, and~the ISAC system could better coordinate the coexistence of radar and communication functions, improving the overall performance and security of vehicular networks~\cite{liang2023}. 
		Song~et~al. proposed a scheme for SL resource allocation in NR-V2X based on 5G cellular mobile communication networks in Mode 1. By~using hybrid spectrum access technology and periodic reporting of channel state information, SL resource allocation was modeled as a mixed binary integer nonlinear programming problem to maximize the total throughput of NR-V2X networks among different subcarriers while complying with total available power and minimum transmission rate constraints. Simulation results showed that the proposed 
		power allocation scheme could save energy, and~the suboptimal SL resource allocation algorithm outperformed other methods
~\cite{Song2020}. 
		MolinaGalan~et~al. conducted an in-depth analysis {and evaluated the} 
		performance of 5G NR-V2X Mode 2 under different traffic patterns
		. The~study pointed out that additional reselections could make SPS more unstable and prone to collisions. Moreover, frequent 
		resource reselections could increase implementation costs. Therefore, {they} 
		proposed suggestions for {an adjusted} 
		reevaluation mechanism to reduce implementation costs and improve system performance. {This work provided valuable insights for the further development and optimization of NR-V2X Mode 2~\cite{molina2023}. }
		Soleymani~et~al. focused on the joint energy efficiency and total rate maximization problem of autonomous resource selection in NR-V2X vehicle communication to meet reliability and latency requirements. They formulated the autonomous resource allocation problem as the ratio of total rate to energy consumption, aiming to maximize the total energy efficiency of power-saving users under reliability and latency requirements. Since the energy efficiency problem is a complex mixed integer programming 
		problem, 
		a traffic-based resource allocation density heuristic algorithm was proposed to address this problem, ensuring the same successful transmission rate as perception-based algorithms while improving energy efficiency by reducing the power consumption\linebreak  per user~\cite{soleymani2021}. 
		
		Hegde~et~al. focused on the efficiency of radio resource allocation and scheduling algorithms in C-V2X communication networks and their impact on latency and reliability. Due to the continuous movement of vehicles, perception-based SPS becomes ineffective, leading to noneffective resource allocation and frequent resource conflicts. Therefore, the~C-V2X communication network was described as a decentralized multi-agent network Markov decision process. Two variants, independent actor-critic and shared experience actor-critic, were proposed, achieving a 15--20\% improvement in reception probability in high vehicle density scenarios~\cite{hegde2023}. 
		Saad~et~al. considered optimizing the medium access control layer in NR-V2X for more effective congestion control. They took into account the AoI indicator in the optimization process and introduced DRL to manage packet transmission rate and transmission power while ensuring high throughput. Compared with traditional distributed congestion control algorithms, the~proposed solution demonstrated better performance in terms of timeliness, throughput, and~average CBR. It highlights the importance and effectiveness of DRL-based congestion control mechanisms in the context of AoI~\cite{saad2023}.
		
		{Currently, NOMA has been applied in many scenarios. Ju~et~al. introduced an energy-efficient sub-channel and power allocation strategy for URLLC-enabled GF-NOMA systems using multi-agent Deep Reinforcement Learning (MADRL). It aims to maximize network energy efficiency while meeting URLLC requirements. It used simulation to check MA2DQN and MADQN performances~\cite{Ju2024NOMA}.
		Tran~et~al. explored secure offloading in vehicular edge computing (VEC) networks with malicious eavesdroppers using NOMA. An~A3C-based scheme was proposed to optimize energy consumption and computation delay. Simulation results demonstrated its advantage in terms of energy efficiency and security~\cite{Tran2023NOMA}.
		Long~et~al. focused on VEC systems, where tasks can be processed locally or offloaded based on vehicle-to-infrastructure (V2I) and vehicle-to-vehicle (V2V) communication. It employed decentralized DRL and the deep deterministic policy gradient (DDPG) algorithm to perform the power allocation while addressing the uncertainty of MIMO-NOMA-based V2I communication and random task arrivals~\cite{long2023power}.
		}
		
		In summary, {the} 
		existing works in the literature on NR-V2X performance analysis and research have not considered the impact of NOMA on AoI{. They also have not considered employing} 
		DRL {in optimizing } 
		AoI and energy consumption in NOMA and NR-V2X based vehicular networks. Therefore, we have undertaken the research presented in this~paper.
		
		\section{System Model and Problem~Formulation}\label{sec3}
		\subsection{Scenario~Description}
\begin{figure}[H]
			\includegraphics[width=5in]{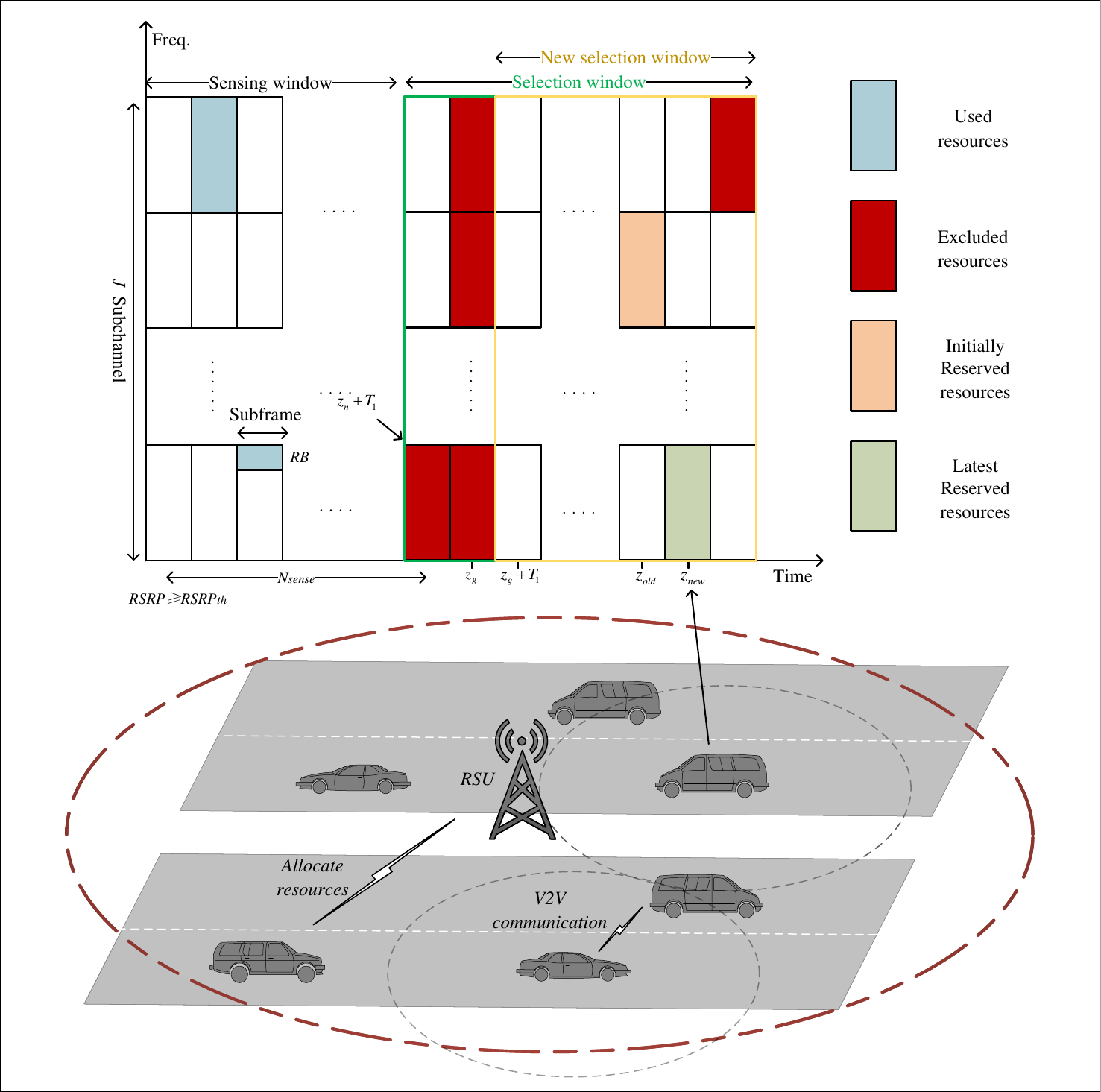}
			\caption{NR-V2X IoV~system.}
			\label{fig1}
		\end{figure}

		{This section describes the AoI and energy optimization model based on NR-V2X Internet of Vehicles (IoV) system as shown in Figure~\ref{fig1}}, focusing primarily on the NR-V2X resource selection method. In~Mode 2, vehicles utilizing NR-V2X 
		adopt a perception-based SPS scheme for dynamic and semi-persistent resource selection. In~the dynamic scheme, resources can only be used once, while in the semi-persistent scheme, resources are reserved for RC times. Additionally, the~re-evaluation mechanism in Mode 2 can detect and avoid potential conflicts in message propagation.
		For NR-V2X Side-Link communication, resources in the time domain are composed of frames and subframes. Each frame typically consists of 10 subframes, {each of which has a duration of} 
		10 ms~\cite{naik2019}. The~subframe is typically 1 ms. In~the frequency domain, the~smallest schedulable frequency unit is the Resource Block (RB). In~the NR-V2X standard, RBs are sequentially combined to form subchannels, allowing vehicles to transmit messages on one or more subchannels~\cite{molina2020}.
		Vehicles continuously monitor channels over a period of time by measuring the Reference Signal Received Power (RSRP) of all $J$ subchannels, storing the latest information of $N_{sense}$ time slots for use as a perception window when resource selection is required.
		{RSRP represents the received power level of the reference signal in a mobile communication system, which is a key indicator for evaluating wireless signal quality and coverage. The~higher the RSRP value, the~stronger the received signal, and the better the signal quality.}
		Then, vehicles initialize a selection window (SW) with a set of consecutive candidate time slots, namely RRI-sized slots. Each vehicle utilizes the information in the perception window to select available communication resources within the SW. Initially, $Z_A$ is set to include all slots in the SW, and~if $z_n$ represents the time slot $n$ following the perception window, then ${Z_A} = \left[ {{z_n} + {T_1},{z_n} + {T_1} + 1, \ldots ,{z_n} + \Gamma } \right]$.
		Then, vehicles exclude all resources corresponding to the subframe from the set $Z_A$ based on certain conditions. Firstly, due to half-duplex communication, vehicles cannot perceive the 
		resources {used} by other vehicles in the same slot 
		in the perception window. Hence, all resources corresponding to the slot need to be excluded from the SW. Secondly, if~the RSRP measurement corresponding to the candidate subframe exceeds RSRP$_{th}$, all resources corresponding to that candidate subframe are excluded from the SW. The~exclusion criterion for the RSRP of the $i$-th subframe (the $j$-th subchannel) in the SW can be expressed as follows:
\begin{equation}
			\begin{aligned}
				RSRP\left( {z_{_{n + T1 + i - {N_{sense}}}}^j} \right) \ge RSR{P_{th}}
			\end{aligned}  .
			\label{RSRP}
		\end{equation}
		If the remaining number of resources in $Z_A$ is less than X\% of the total available resources, RSRP$_{th}$ is increased by 3 dB
		. In~NR-V2X Mode 2, X can be set to 20, 35, or~50. Finally, vehicles randomly select a communication resource from the remaining resources in $Z_A$ to reserve for subsequent transmission use, transmitting RC times at intervals of RRI. RC varies with RRI to ensure the 
		time of the selected resource is between 0.5 and 1.5 s. {Therefore, as~shown in reference~\cite{Ali2021NR}, the~initial value of RC for a vehicle RC0 is \mbox{represented as follows:}}
\begin{equation}
			\begin{aligned}
				RC_i^0 = \left\{ {\begin{array}{*{20}{c}}
						{\frac{{1000}}{{\max (20,\Gamma )}}}&{20 \le \Gamma  \le 100}\\
						{50}&{1 \le \Gamma  \le 19}
				\end{array}} \right.
			\end{aligned}  .
			\label{RC}
		\end{equation}
		
		After \hl{RC} 
 decreases to 0, vehicles {can continue to use }
		the preselected resources with a probability of $P_{rk}$ or reselect new resources for transmission with a probability of $1-P_{rk}$. Before~transmitting messages,vehicles that select communication resources in time slot $z_{old}$ may check whether these resources are still available using a reassessment mechanism (i.e., not reserved by another vehicle) \cite{dayal2023}. Vehicles will perform the reassessment check in time slot $z_g$. The~new resource selection window, denoted as SW', is defined as $\left[ {{z_g} + {T_1}, \ldots ,{z_n} + \Gamma } \right]$. If~resources previously excluded are found to be available again during the reselection process, vehicles will select new resources from the available resources in SW'. The~resources initially chosen in time slot $z_{old}$ will be replaced by new resources {in}
		time slot $z_{new}$ as depicted in the figure. Table~\ref{tab1} lists the parameters used in this~chapter.
		
		\begin{table}[H]\renewcommand{\arraystretch}{1.2}
			\footnotesize
			\caption{The summary for~notations.}
			\label{tab1}
		\setlength{\tabcolsep}{6.25mm}

\begin{adjustwidth}{-\extralength}{0cm}
	\begin{tabular}{cm{5cm}cm{5cm}}
				\toprule
				\textbf{Notation} &\textbf{Description} &\textbf{Notation} &\textbf{Description}\\
				\midrule
				{$N_v$} &{Total number of vehicles.} &$D$ &\multicolumn{1}{m{6cm}}{Road length.}\\
				\midrule
				{$D_{RSU}$} &{RSU coverage range.} &$w$ &\multicolumn{1}{m{6cm}}{The maximum distance of the vehicle as the receiver.}\\
				\midrule
				$RC$ &The value of the resource counter. &$\Gamma$ &\multicolumn{1}{m{6cm}}{The size of SW and RRI.}\\
				\midrule
				$\eta $ &SINR. &${h_s}$ &\multicolumn{1}{m{6cm}}{Small scale fading gain.}\\
				\midrule
				$h$ &Large scale fading gain. &$p$ &\multicolumn{1}{m{6cm}}{Vehicle transmission power.} \\
				\midrule
				$L_d$ &{Path loss.} &$d$ &\multicolumn{1}{m{6cm}}{Distance between communication vehicles.}\\
				\midrule
				$I$ &Interference signal power. &$p_n$ &\multicolumn{1}{m{6cm}}{noise power.}\\
				\midrule
				$\sigma $ &Signal overlap degree. &$C$ &\multicolumn{1}{m{6cm}}{Received signal power.}\\
				\midrule
				{$u$} &{Communication situation}. &$W$ &\multicolumn{1}{m{6cm}}{Resource channel bandwidth for vehicle selection.}\\
				\midrule
				$G$ &Message size. &$\Phi $ &\multicolumn{1}{m{6cm}}{The AoI at the receivers.}\\
				\midrule
				$\varphi$ &The AoI of messages in the queue. &$\beta$ &\multicolumn{1}{m{6cm}}{Has vehicle transmitted messages.}\\
				\midrule
				$q$ &Queue length. &$L$ &\multicolumn{1}{m{6cm}}{Queue capacity.}\\
				\midrule
				$E$ &Energy consumption generated by vehicles in reserved resources. &${l_i}$ &\multicolumn{1}{m{6cm}}{Communication time of vehicle i.}\\
				\midrule
				$N$ &{Number of receivers for a certain vehicle.} &{$P_{}^t(u_{}^t = 1)$} &\multicolumn{1}{m{6cm}}{{The probability of the receiving end successfully receiving the message.}}\\
				\midrule
				${\theta _x}$ &The weights of the transition function strategy network between discrete and continuous actions. &${\theta _Q}$ &\multicolumn{1}{m{6cm}}{The weights of state action value function strategy networks.}\\
				\midrule
				$l{r_x}$ &Learning rate of transition function strategy networks between discrete and continuous actions. &$l{r_Q}$ &\multicolumn{1}{m{6cm}}{Learning rate of state action value function strategy network.}\\
				\bottomrule
			\end{tabular}
\end{adjustwidth}
		\end{table}
\unskip

		\subsection{NR-V2X-NOMA Communication~Model}
		In the NR-V2X vehicular networking system, {we denote the transmitter as $i$, receiver vehicle as $j$}, 
		and the considered time slot as $t$. SINR is represented as follows:
\begin{equation}
			\begin{aligned}
				\eta _{i \to j}^t = \frac{{h_s^t{{h_{i \to j}^tp_i^t} \mathord{\left/
				{\vphantom {{h_{i \to j}^tp_i^t} {{L_d}(d_{i \to j}^{})}}} \right.
				\kern-\nulldelimiterspace} {{L_d}(d_{i \to j}^{})}}}}{{I_{i \to j}^t + p_n^{}}}
			\end{aligned}  ,
			\label{SINR1}
		\end{equation}
		where $p_i^t$ is the transmission power by vehicle $i$, $h_s^t$ is the random small-scale fading gain, $h_{i \to j}^t$ is the large-scale fading gain of the link from $i$ to $j$ in time slot $t$, $L(d_{i \to j}^{})$ is the path loss as a function of the distance from $i$ to $j$, $p_n$ is the noise power, and~$I_{i \to j}^t$ is the interference power. In~Equation~(\ref{SINR1}), the~numerator represents the 
		received power, while the denominator is the sum of the noise power and interference, assumed to be Gaussian with zero mean. $I_{i \to j}^t$ is defined as follows:
\begin{equation}
			\begin{aligned}
				I_{i \to j}^t = \sum\limits_{k \in {V^t},k \ne i} {\sigma _{k,i}^{t}} \frac{{h_s^th_{k \to j}^tp_k^t}}{{{L_d}(d_{k \to j}^{})}}
			\end{aligned}  ,
			\label{SINRI}
		\end{equation}
		where $V^t$ is the set of nodes transmitting in time slot $t$, $\sigma _{k,i}^t$ is a multiplicative coefficient between 0 and 1 that quantifies the interference power of $k$ in the subchannel used by $i$, relative to the transmission power of $k$. If~$k$ uses the exact same subchannel as $i$, $\sigma _{k,i}^t$ is 1; if its signal does not overlap or only partially overlaps, $\sigma _{k,i}^t$ is less than 1. The~calculation of $\sigma _{k,i}^t$ takes into account the in-band emission, consistent with the specifications in~\cite{lin2019}.
		
		The receiver $j$ employs a serial interference cancellation mechanism in NOMA to decode multiple messages in different subchannels. The~decoding process involves selecting the message with the maximum power as the desired signal, while treating the others as interference signals
		. If~we denote the signal $i$ with the current maximum received power {as} follows:
\begin{equation}
			\begin{aligned}
				C_{i \to j}^t = \frac{{h_s^th_{i \to j}^tp_i^t}}{{{L_d}(d_{i \to j}^{})}}
			\end{aligned}  ,
			\label{SINRCi}
		\end{equation}
		{then,} 
		other signals with received power lower than signal $i$ are denoted as follows:
\begin{equation}
			\begin{aligned}
				C_{k \to j}^t = \frac{{h_s^th_{k \to j}^tp_k^t}}{{{L_d}(d_{k \to j}^{})}}
			\end{aligned}  .
			\label{SINRCk}
		\end{equation}

		The set of other vehicles whose received signal power is {lower} 
		than vehicle $i$ is denoted as ${{\cal I}_i} = \{ k \in {V_i}\backslash i\mid C_k^t < C_i^t\} $. Then, the~expression for the SINR obtained by vehicle $j$ using NOMA is as follows:
\begin{equation}
			\begin{aligned}
				\eta _{i \to j}^t = \frac{{C_{i \to j}^t}}{{\sum\limits_{k \in {{\cal I}_i}} {C_{k \to j}^t\sigma _{k,i}^{t}}  + p_n^{}}}
			\end{aligned}  .
			\label{SINRNOMA}
		\end{equation}	
			
		{When vehicle j decodes the message with the highest power from vehicle i, the~message power of the vehicle in ${{\cal I}_i}$ is used as the interference power, and~the magnitude of the interference is affected by the degree of channel overlap $\sigma$. Therefore, by~adjusting the power allocation, the~SINR of each message can be increased, thereby increasing the likelihood of successful communication.}

		\subsection{AoI~Model}
		When the size of the transmitted message is $G$, the~criterion for successful communication between vehicles $i$ and $j$ is as follows:
\begin{equation}
			\begin{aligned}
				u_{i \to j}^t = \left\lfloor {\frac{{W_i^t{{\log }_2}(1 + \eta _{i \to j}^t)}}{G}} \right\rfloor 
			\end{aligned}  ,
			\label{u}
		\end{equation}
		where {$\lfloor\rfloor$ indicates rounding down the values in it, ${{\log }_2}(1 + \eta _{i \to j}^t)$ represents transmission rate,} and $W_i^t$ represents the bandwidth utilized by vehicle $i$ for message transmission. {So, }
		$u_{i \to j}^t = 0$ indicates that the communication rate between vehicles $i$ and $j$ is insufficient to transmit the message within the specified time slot $t$, resulting in communication failure. Due to the nature of NR-V2X, where each transmission between vehicles requires waiting for a time equivalent to the RRI, each failed transmission between vehicles leads to an increase in the AoI at the receiving end vehicle. Successful transmission, on~the other hand, results in an increase {in} 
		AoI at the receiving vehicle by $\Gamma$ 
		. The~change in AoI at the receiving end vehicle $j$ between communication time slots can be expressed as follows:
\begin{equation}
			\begin{aligned}
				\Phi_{i \rightarrow j}^{t+\Gamma}=\left\{\begin{array}{cc}
					\varphi_{i n}^{t ,1}+\Gamma & u_{i \rightarrow j}^{t}=1 \\
					\Phi_{i \rightarrow j}^{t}+\Gamma & u_{i \rightarrow j}^{t}=0
				\end{array}\right.
			\end{aligned}  .
			\label{Phi}
		\end{equation}

		It can be observed that the AoI at the receiving end is influenced by the transmission interval size $\Gamma$, 
		transmission status $u$, and~the 
		AoI of the message transmitted by vehicle $i$, where $\Phi _{i \to i}^{} = 0$. Among~these factors, when the transmission interval $\Gamma$ is smaller, the~receiving end has more opportunities to update to the AoI of the transmission end. Additionally, a~higher transmission success rate increases the likelihood of updating to the AoI at the transmission end. Furthermore, the~AoI at the transmission end decreases as the queue processing rate $\frac{1}{\Gamma} $ increases. Therefore, when $\Gamma$ is smaller, the~AoI at the transmission end is smaller, resulting in a smaller AoI at the receiving end as~well.
		
		The average AoI at the receiving end for all vehicles in the system is defined as follows:
\begin{equation}
			\begin{aligned}
				\overline \Phi   = \frac{1}{T}\frac{1}{{N_v^2}}\sum\limits_{i \in {N_v}} {\sum\limits_{j \in {N_v}}^{} {\Phi _{i \to j}^{}} }  
			\end{aligned}  ,
			\label{avgPhi}
		\end{equation}
		
		In the scenario where four message types are considered, the $\varphi_{i n}^{t+1,b}$ is represented as follows:
\begin{equation}
			\begin{aligned}
				\varphi _{i,n}^{t + 1,b} = \beta _{i,n}^t\left( {\varphi _{i,n}^{t,b + 1} - \varphi _{i,n}^{t,b}} \right) + \varphi _{i,n}^{t,b} + 1  
			\end{aligned}  ,
			\label{phi}
		\end{equation}
		
		In Equation~(\ref{phi}), $n$ represents multiple message types, $b$ denotes the position of the message in the queue, and $\beta _{i,n}^t \in \left\{ {0,1} \right\}$ indicates whether a type $n$ message can be transmitted (with 1 indicating it can be transmitted), where the queue operates on a first-in-first-out basis. And~the parameter $\Gamma$ determines the frequency of $\beta _{i,n}^t = 1$, so $\beta _{i,n}^t $ can be \mbox{expressed as follows:}
\begin{equation}
			\begin{aligned}
				\beta _{i,n}^t = \left\lfloor {\frac{t}{{{z_{new}} + m\Gamma }}} \right\rfloor \left\lceil {\frac{{q_{i,n}^t}}{L}} \right\rceil  
			\end{aligned}  .
			\label{beta}
		\end{equation}
		where {$\lceil\rceil$ represents rounding up the values in it}, $z_{new}$ represents the time slot allocated for vehicle reservation in the SW, $m$ indicates the number of times a vehicle uses reserved resources, and $q$ and $L$ represent the queue length and queue capacity, respectively. Due to the consideration of multiple priority queues in this scenario, when the high-priority message is in queue $\beta _{i,n}^t = 1$, the~$\beta _{i,n}^t = 1$ of other queues~applies.
			
		\subsection{Energy Consumption~Model}
		In NR-V2X, when vehicle $i$ reselects resources, 
		the energy {consumption for the }
		previously reserved resources is given by the following:
\begin{equation}
			\begin{aligned}
				E_i^t = p_i^tl_i^tRC_i^0
			\end{aligned} .
		\end{equation}
		
		$p_i^t{l_i}$ represents the energy consumption generated by using reserved resources at a time, and~$l_i$ represents the {time when }
		vehicle \( i \) utilizes resources, which is the size of one time slot. As~shown in Equation~(\ref{RC}), when the transmission interval is smaller, $RC^0_i$ is larger, and~the energy consumption during this time period is greater, indicating a trade-off between energy consumption and~AoI.
		
		The average energy consumption of all vehicles in the system is defined as
\begin{equation}
			\begin{aligned}
				\bar E = \frac{1}{T}\frac{1}{{{N_v}}}\sum\limits_{t \in {\cal T}} {\sum\limits_{i \in {N_v}} {E_i^t} } 
			\end{aligned}  .
			\label{eq12}
		\end{equation}

		\section{Optimization Method of MPDQN Based on~DRL}\label{sec4}
		\subsection{Framework for Optimization~Problems}
		Based on the defined system model, the~optimization problem is formulated to minimize the weighted sum of the average AoI and energy consumption of vehicles in the system. Since the AoI and energy consumption of vehicles depend on RRI $\Gamma$ and power $p$, the~optimization problem can be {expressed} 
		as follows:
\begin{align}
			\label{eqmin}
			&\min_{{\Gamma}^t,{p}^t} \left[\omega_1\overline{E}+\omega_2\overline\Phi \right]\\
			s.t.\quad&p^{t}\in [0,P_{max}], \forall t \in \mathcal{T}, \tag{\hl{15{a}}
} \label{eqmina}\\
			&\Gamma^{t}\in\{20,50,100\}, \forall t \in \mathcal{T}, \tag{\hl{15{b}}} \label{eqminb}
		\end{align}
		{where} $\omega_1$ and $\omega_2$ are non-negative weight~factors.
		
		Since the channel conditions in the NR-V2X system are uncertain, we employ the Multi-Pass deep Q-Networks (MPDQN) method based on DRL to solve this optimization problem. In~this method, the~RSU serves as the agent, and~its observed state at time slot $t$ is as follows:
\begin{equation}
			\begin{aligned}
				{S^t} = \left[ {s_1^t, \cdots ,s_i^t, \cdots ,s_{{N_v}}^t} \right] 
			\end{aligned}  .
			\label{eq12}
		\end{equation}
		
		The state of each vehicle is defined as follows:
\begin{equation}
			\begin{aligned}
				s_{}^t = \left[ {N_{}^t,\bar d_{}^t,P_{}^t(u_{}^t = 1),RC_{}^0} \right] 
			\end{aligned}  ,
		\end{equation}
		where $N^t$ is the total number of other vehicles within the range $w$, defined as receivers. The~average distance to these receivers is $\bar d^t$. $P_{}^t(u_{}^t = 1)$ is the probability of successful message reception by the receivers. $RC^0$ represents the total number of times {that} vehicles use reserved~resources.
		
		The action assigned by the RSU to vehicles at the time slot $t$ is as follows:
\begin{equation}
			\begin{aligned}
				a_{}^t = \left( {\Gamma _{}^t,p_\Gamma ^{}} \right) 
			\end{aligned}  ,
		\end{equation}
		where $p_\Gamma ^{}$ represents the parameter that converts the continuous action \( p \) into a discrete action \(\Gamma \). They act as two sub-actions, and~the tuple they form constitutes the complete action~assigned.
		
		The objective of the optimization problem is to minimize the AoI and energy consumption in the system. Therefore, the~reward function is defined as follows:
\begin{equation}
			\begin{aligned}
				r_i^t =  - \left( {{\omega _1}E_i^t + {\omega _2}\overline {\Phi _i^t} } \right) 
			\end{aligned}  .
		\end{equation}
		where $\overline {\Phi _i^t}$ is the mean AoI of the receivers for vehicle \( i \) over a certain period of time:
\begin{equation}
			\begin{aligned}
				\overline {\Phi _i^t}  = \frac{1}{T}\frac{1}{{{N_v}}}\sum\limits_{t = 1}^T {\sum\limits_{j \in {N_v}}^{} {\Phi _{i \to j}^t} } 
			\end{aligned}  .
		\end{equation}
		
		\subsection{Solution to Optimization~Problems}

\begin{figure}[H]
			\includegraphics[scale=0.5]{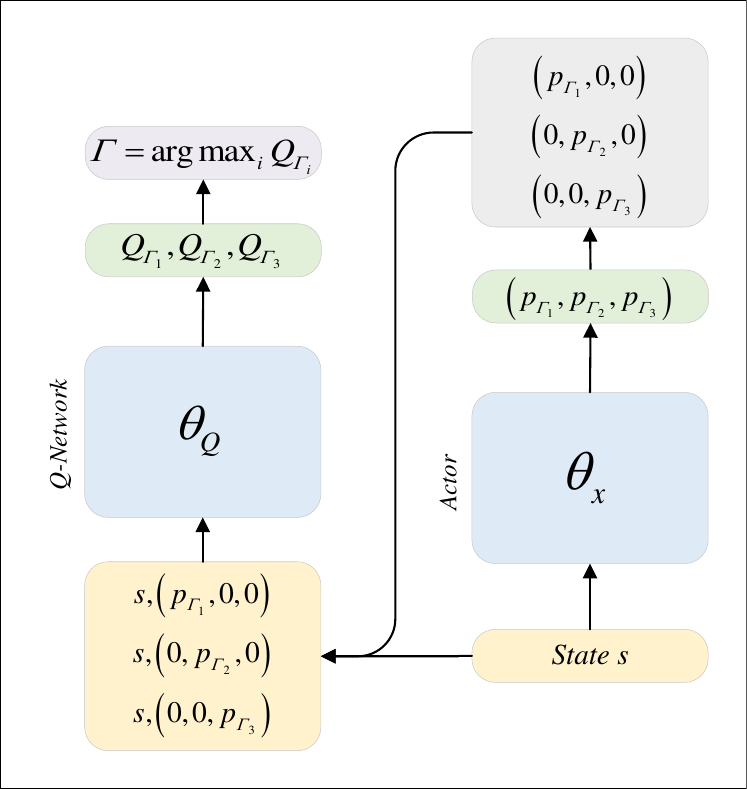}
			\caption{The process of agent selecting actions in~MPDQN.}
			\label{MPDQN}
		\end{figure}

		For action tuple $\left( {\Gamma ,{p_\Gamma }} \right)$, a~policy network is used to match them.
\begin{equation}
			p_\Gamma ^t = {x^Q}\left( {{s^t},\Gamma ;{\theta _x}} \right),
		\end{equation}
		Where ${\theta _x}$ represents the weights of this network. Then, another deep neural network is used to approximate the action-value function $Q\left( {s,\left( {\Gamma ,x} \right)} \right)${:}
\begin{equation}
			Q(s_{}^t,a_{}^t) = Q\left( {s_{}^t,\left( {\Gamma _{}^t,{x^Q}\left( {s_{}^t,\Gamma ;{\theta _x}} \right)} \right);{\theta _Q}} \right),
		\end{equation}
		where the network weights are denoted as ${\theta _Q}$. The~process {where} 
		the agent obtains the action with the highest action value is illustrated in Figure \ref{MPDQN}.
		
		The loss functions for the  Q-Network and the network of the actor  are defined as follows:
\begin{equation}
			\begin{aligned}
				\begin{array}{l}
					L_{Q}\left(\theta_{Q}\right)= 
					\underset{\left(s^t, (\Gamma, p_{\Gamma}), r^t, s^{t+1}\right) \sim M}{\mathbb{E}}\left[\frac{1}{2}\left(y^{t}-Q\left(s^t,\left(\Gamma, p_{\Gamma}\right) ; \theta_{Q}\right)\right)^{2}\right]
				\end{array}
			\end{aligned}  ,
			\label{LossQ}
		\end{equation}
\begin{equation}
			\begin{aligned}
				\begin{array}{l}
					L_{x}\left(\theta_{x}\right)= 
					\underset{s^t \sim M}{\mathbb{E}}\left[- \underset{{\Gamma=\{20,50,10\}}}\sum Q\left(s^t, \Gamma, x^{Q}\left(s, \Gamma ; \theta_{x}\right) ; \theta_{Q}\right)\right]
				\end{array}
			\end{aligned} ,
			\label{Lossx} 
		\end{equation}
		where \( y_t \) is defined as follows:
\begin{equation}
			y_{}^t = r_{}^t + \gamma \mathop {\max }\limits_\Gamma  Q\left( {{s^{t + 1}},{\Gamma ^{t + 1}},{x^Q}\left( {{s^{t + 1}};{\theta _x}} \right);{\theta _Q}} \right).
		\end{equation}
		
		Finally, the~network weights are updated using the learning rates $l{r_x}$ and $l{r_Q}$ for each network, aiming to approach the optimization objective.
\begin{equation}
			\begin{aligned}
				\theta _Q^{t + 1} = \theta _Q^t - l{r_Q}{\nabla _{{\theta _Q}}}{L_Q}\left( {{\theta _Q}} \right)
			\end{aligned}  ,
			\label{updateQ}
		\end{equation}
\begin{equation}
			\begin{aligned}
				\theta _x^{t + 1} = \theta _x^t - l{r_x}{\nabla _{{\theta _x}}}{L_x}\left( {{\theta _x}} \right)
			\end{aligned}  ,
			\label{updatex}
		\end{equation}
		
		Next, we will describe the algorithm process in detail. First, the~parameters of both networks are randomly initialized, and~an experience replay buffer of size \( M \) is established. Then, the~algorithm iterates over \( EP \) episodes. At~the beginning of each episode, the~system 
		parameters are reset. The~RSU selects initial action tuples based on the initial state and the network and~observes the next state. Subsequently, the~algorithm iterates from timeslot 1 to timeslot \( T \). For~each timeslot \( t \), the~RSU allocates actions to vehicles needing resource reallocation based on the current state. When selecting actions, the~RSU either explores randomly with a certain probability or chooses the action with the maximum Q-value, introducing exploration noise to avoid local optima. Finally, the~tuple $\left[ {{s^t},\left( {{\Gamma ^t},{p_\Gamma }} \right),{r^t},{s^{t + 1}}} \right]$ is stored in the experience replay buffer. When the number of tuples in the buffer exceeds the sample size \( B \), {they will be employed }
		to update the network parameters. The~pseudocode is shown in Algorithm \ref{al1}.
		\setlength{\algoheightrule}{1pt} 

\setlength{\algotitleheightrule}{0.5pt}

		\begin{algorithm}[H]
			\caption{Optimization algorithm for AoI and energy consumption based on~MPDQN}
			\label{al1}
			\KwIn{$\gamma$, $l{r_x}$, $l{r_Q}$}
			\KwOut{optimized $\theta_Q$, $\theta_x$}
			Initialize the learning rates $l{r_x}$ and $l{r_Q}$, experience replay buffer $ M $, sample size $ B $, network weights $\theta_x$ and $\theta_Q$.\\
			\For{episode from $1$ to $EP$ }
			{
				Reset the model parameters;\\
				Observe the initial state $s^0$ and output the initial action $\left( {\Gamma _i^0,{p_\Gamma }} \right)$;\\
				Obtain the state $s_i^1 = \left[ {N_i^1,d_i^1,Rn_i^1,RC_i^0} \right]$.\\
				\For{slot $t$ from $1$ to $T$ }
				{
					Obtain action tuples based on the policy network or through random exploration:\\
					$a_{}^t = \left\{ {\begin{array}{*{20}{c}}
							{random\left( {{\Gamma ^t},{p_\Gamma }} \right) + \Delta }&{{P_{ran}}}\\
							{\mathop {\arg \max }\limits_\Gamma  Q\left( {s_{}^t,{\Gamma ^t},x;{\theta _Q}} \right) + \Delta }&{1 - {P_{ran}}}
					\end{array}} \right.$\\
					Execute action $\left( {\Gamma _i^t,{p_\Gamma }} \right)$, observe state $\boldsymbol{s}^{t+1}$ and reward $r^t$\;
					Store transition tuple $\left[ {{s^t},\left( {{\Gamma ^t},{p_\Gamma }} \right),{r^t},{s^{t + 1}}} \right]$ to $M$\;
					\If {number of tuples in $M$ is larger than $B$ }
					{
						Randomly sample $B$ transitions tuples from $M$\;
						Iterate networks according to Equation \eqref{LossQ},  \eqref{Lossx},  \eqref{updateQ} and  \eqref{updatex}.
					}
				}
			}
		\end{algorithm}
		\vspace{+9pt}
		{During the testing phase, there is no need to update the parameters. The actions are assigned to the required vehicles based on the optimized strategy in the training phase to carry out the test. The~corresponding pseudocode is shown in Algorithm \ref{al2}.}
		\vspace{+9pt}
		
		\begin{algorithm}[H]
			\caption{Testing stage of the~MPDQN}
			\label{al2}
			\For{episode from $1$ to $EP$ }
			{
				Reset the model parameters;\\
				Receive initial observation state $\boldsymbol{s}^{1}$;\\
				\For{slot $t$ from $1$ to $T$ }
				{
					Assign RRI and power to vehicles that need to occupy communication resources according to the policy and their status.\\
					Execute action $\left( {\Gamma _i^t,{p_\Gamma }} \right)$\\
					Observe state $\boldsymbol{s}^{t+1}$ and reward $r^t$\;
				}
			}
		\end{algorithm}
		
		\section{Simulation Results and~Analysis}\label{sec5}
		\subsection{Parameter~Settings}
		The simulation {has been }
		conducted using Python 3.6 and MATLAB 2023b, based on modifications and simulations built upon the code provided in~\cite{todisco2021}. The~simulation scenario involves a two-way highway covered by the RSU communication range, where randomly distributed vehicles travel at
		constant speeds in their respective lanes and use NR-V2X Side-Link technology for V2V communication. The~length of the highway, $D$, is 500 m, the~RSU coverage range, $D_{RSU}$, is 250 m, and~the maximum distance, $w$, between~vehicles and receivers is 150 m. All vehicles utilize four different priority queues of length $L$.	
		{And the vehicles occupy a channel bandwidth of 10 MHz within the 5.9 GHz frequency band. The~receiver has a noise figure of 9 dB. 
			 The~path loss model features a standard deviation of 3 dB and a decorrelation distance of 25 m. Thus, the~$RSRP_{th}$ is $-$126 dBm.}
		MPDQN employs a neural network with one hidden layer and updates its parameters using the Adam optimization method with learning \hl{rates} 
 $l{r_Q} = 5~\times~{10^{ - 4}}$ and $l{r_x} = {10^{ - 4}}$. The~experience replay buffer size $M$ is set to 2000, and~the sample size $B$ is 128~\cite{kingma2014}. Ornstein--Uhlenbeck noise is used as the exploration noise for the network, with~a decay rate set to 0.15 and variance set to 0.0001. {The key simulation parameters have been listed in }
	 Table~\ref{tab2}.
		
		\begin{table}[H]
			\caption{Values of the parameters in the~experiments.}
			\label{tab2}
		\setlength{\tabcolsep}{9.2mm}

			\begin{tabular}{cccc}
				\toprule
				\textbf{Parameter} &\textbf{Value} &\textbf{Parameter} &\textbf{Value}\\
				\midrule
				$N_v$ &$20$, $30$, $40$, $50$ &$L$ &10\\
				\midrule
				$D$ &$500$ m &$D_{RSU}$ & $250$ m\\
				\midrule
				$w$ &$150 $ m &$P_{max}$ &$23 $ dBm\\
				\midrule
				$v_{min}$ & $60 $ km/h &$v_{max}$ & $80 $ km/h\\
				\midrule
				{$W$} & {$10 $} MHz&{$RSRP_{th}$} & {$-126 $} dB\\
				\midrule
				$lr_Q$ &$5~\times~10^{-4}$ &$lr_{x}$ &$10^{-4}$\\
				\midrule
				$\tau$ &$0.01$ &$\gamma$ &0.99\\
				\midrule
				$M$ &$2000$ &$B$ &$128$ \\
				\bottomrule
				
			\end{tabular}
		\end{table}
\unskip
		
		\subsection{Simulation~Result}
		In this section, we first compare the AoI of vehicles in LTE-V2X and NR-V2X. Then, we compare the AoI of vehicles in NR-V2X before and after using NOMA, as~well as the AoI in LTE-V2X and NR-V2X based on NOMA. Finally, we optimize the joint objective of AoI and energy consumption in NR-V2X using MPDQN. 
		{Many recent works have employed genetic algorithms~\cite{bulut2020energy} and random algorithms~\cite{Ullah2022NOMA} as baseline algorithms for resource allocation, and~thus, we shall compare our approach with these two methods above.}	

\begin{figure}[H]
			\includegraphics[scale=0.5]{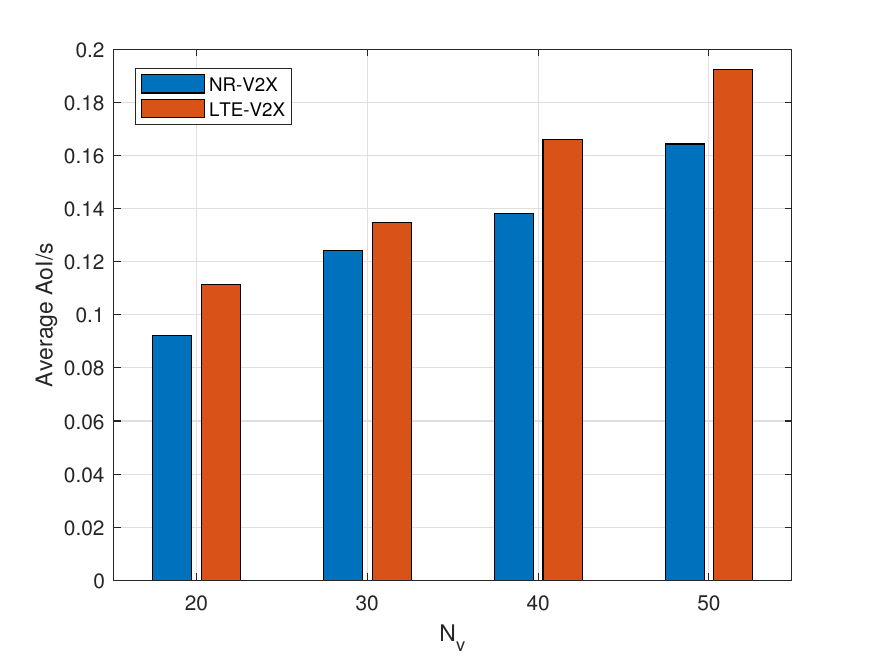}
			\caption{Average AoI between LTE-V2X and~NR-V2X.}
			\label{OMA}
		\end{figure}

	 Figure~\ref{OMA} illustrates the variation in average AoI in the system as the number of vehicles using LTE-V2X and NR-V2X for V2V direct communication scenarios. The~number of vehicles considered are 20, 30, 40, and~50, with~each vehicle following 3GPP standards and employing a random strategy to select its RRI (Resource Reuse Interval) and transmission power. It is observed that the average AoI in the system increases with the number of vehicles, regardless of whether the LTE-V2X or NR-V2X communication mode is utilized. This AoI increase can be attributed to the expansion of the receiver set within the communication range of vehicles as the number of vehicles in the system increases, leading to increased interference among them. Furthermore, due to the half-duplex communication mode, more vehicles are unable to receive messages due to resource contention, resulting in an increase in the average AoI. Additionally, as~NR-V2X serves as a complement and advancement to LTE-V2X, the~average AoI of vehicles in a vehicular networking system using NR-V2X for communication operations is consistently lower than that in an LTE-V2X~system.
	 
\begin{figure}[H]
			\includegraphics[scale=0.5]{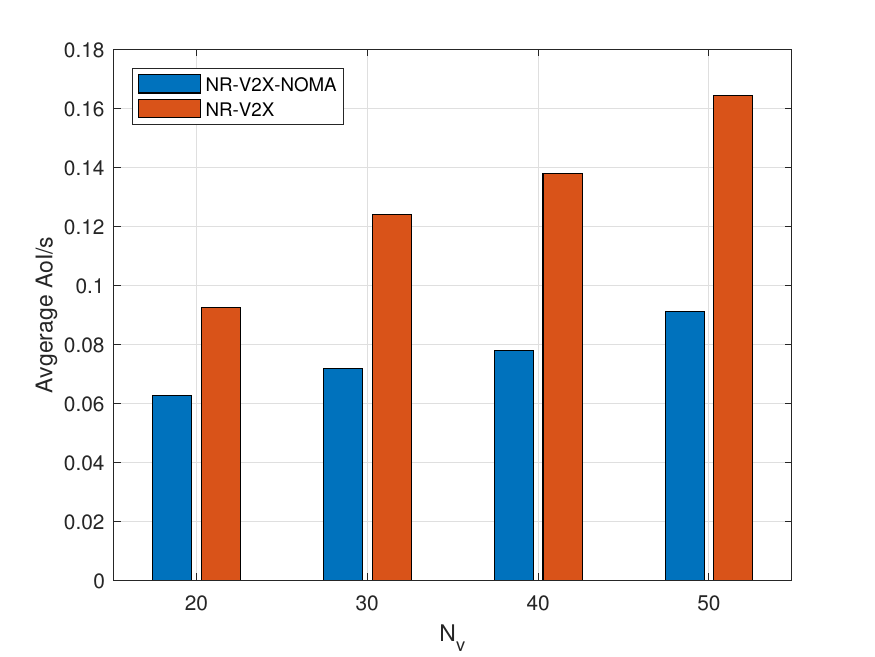}
			\caption{Average AoI before and after using NOMA for~NR-V2X.}
			\label{N-O}
		\end{figure}

	 Figure~\ref{N-O} illustrates the variation in average AoI in the system as the number of vehicles changes when vehicles use NR-V2X and V2V communication with NOMA enabled. In~this scenario, vehicles randomly select their RRI (Resource Reuse Interval) and transmission power.
		It can be observed that when vehicles use NOMA, they exhibit lower AoI {when the number of vehicles changes}
		. Moreover, the~overall growth trend is smoother. This is attributed to NOMA's power-domain-based decoding approach, which significantly mitigates the impact of different vehicles occupying the same resources.
		Similarly, the~average AoI within the system increases with the number of vehicles due to the proliferation of receivers. When more receivers cannot successfully receive messages due to resource contention, the~AoI increases as the number of vehicles become~larger.

	 \vspace{-12pt}

\begin{figure}[H]
			\includegraphics[scale=0.5]{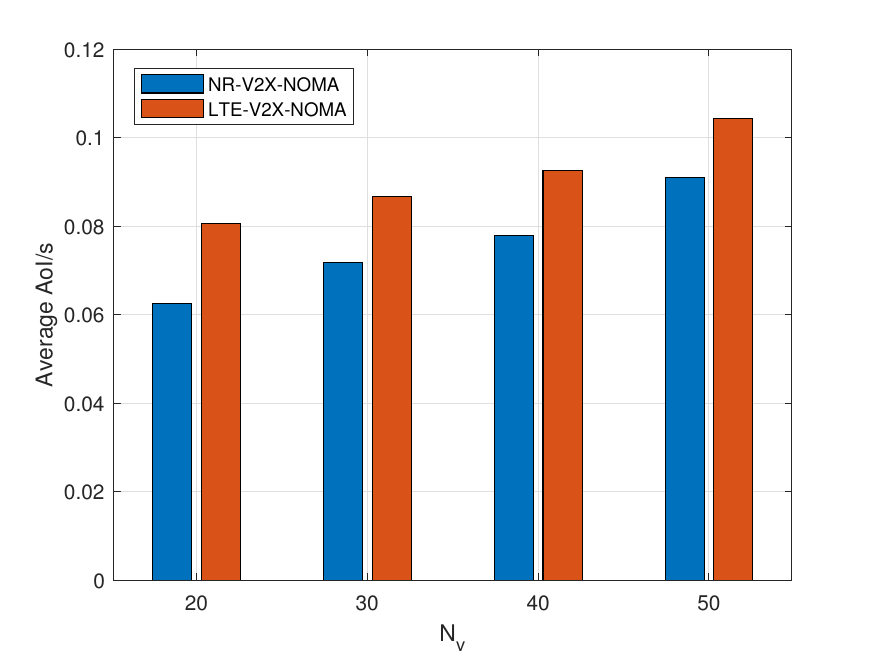}
			\caption{Average AoI between LTE-V2X and NR-V2X based on~NOMA.}
			\label{NOMA}
		\end{figure}

	 Figure~\ref{NOMA} depicts the variation in average AoI within the system as the number of vehicles changes when vehicles utilize NOMA-based NR-V2X and LTE-V2X for V2V communication. With~NOMA incorporated, the~relationship between the two remains consistent with Figure~\ref{OMA}, where the average AoI of NR-V2X consistently remains lower than that of LTE-V2X scenarios.
		This persistent advantage of NR-V2X over LTE-V2X can be attributed to the fact that vehicles in NR-V2X experience fewer resource collisions even before the implementation of~NOMA.

\vspace{-6pt}

\begin{figure}[H]
		\hspace{-9pt}	\includegraphics[scale=0.5]{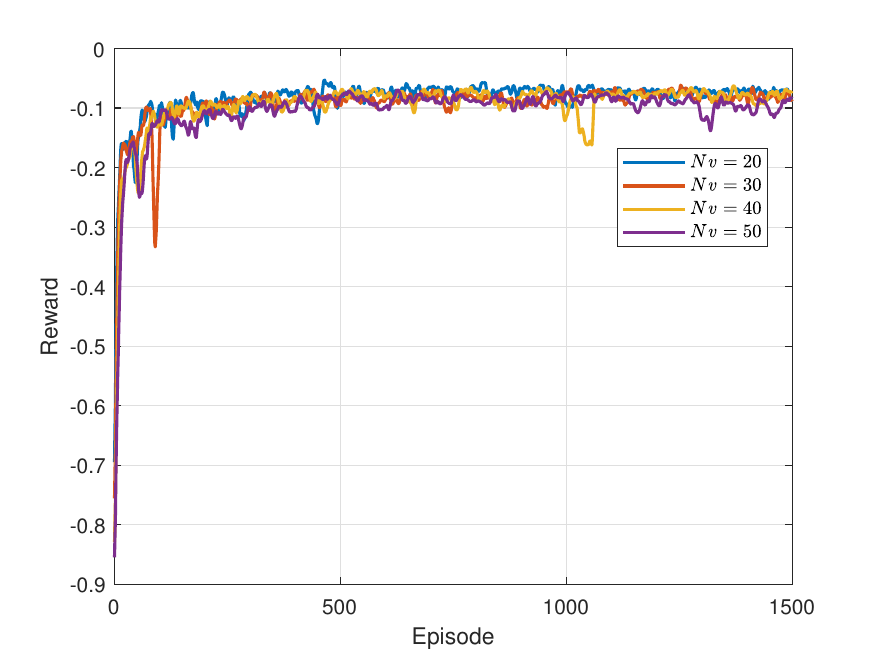}
			\caption{\hl{Learning} 
 curves under different numbers of~vehicles.}
			\label{reward}
		\end{figure}

	 Figure~\ref{reward} illustrates the learning curves of training under different scenarios. 
		Overall, it can be observed that the rewards of different curves exhibit an upward trend in fluctuation from episode 0 to episode 500, 
		Subsequently, the~learning curves stabilize, indicating that the agent has learned a strategy close to optimal.
		There is some jitter in the curves around episode 1000 when the number of vehicles is 40 and 50. This is attributed to detection noise impacting the agent, necessitating adjustments to return to a convergent state.
		Furthermore, it is noticeable that as the number of vehicles increases, the~rewards decrease. This is due to the increasing interference experienced by each device with the growing number of devices in the system, resulting in lower SINR. This leads to prolonged transmission delays 
		and 
		increases {the} system AoI. To~maintain a lower AoI, RSUs notify more vehicles to utilize communication resources for transmission, that is, more vehicles imply a higher AoI and increased energy consumption, and hence, lower~rewards.

\begin{figure}[H]
			\includegraphics[scale=0.5]{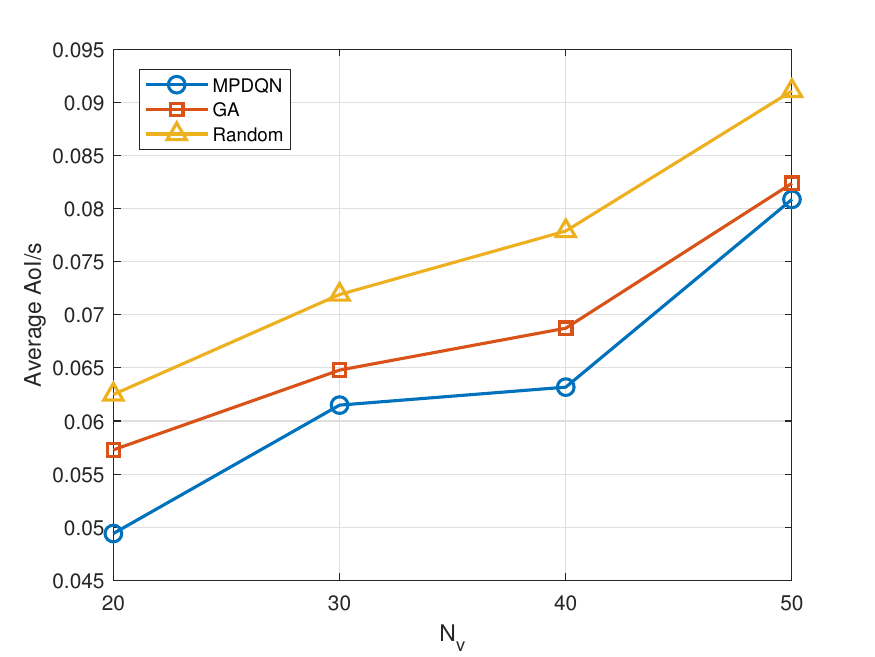}
			\caption{Average AoI with $N_v$ under different~algorithms.}
			\label{RhoAoI}
		\end{figure}

	 Figure~\ref{RhoAoI} depicts the variation in average AoI {with respect to different }
		numbers of vehicles in the NOMA-based NR-V2X vehicular network system when employing MPDQN, genetic algorithm, and~random algorithm strategies. It can be observed that the AoI of all three strategies increases {when the number of devices increases}
		. This is attributed to the interference experienced by each device as the number of devices increases, leading to increased transmission times according to the equation, which may further increase the AoI of systems.
		Furthermore, the~allocation strategies obtained by MPDQN, which approximates the optimal strategy, and the ~genetic algorithm consistently outperform the random strategy. This is because the near-optimal strategy obtained by MPDQN selects actions for vehicles based on observed states, while the genetic algorithm derives better action allocation strategies through evolution, whereas the random strategy merely generates action allocations randomly.
		Additionally, it can be observed that the strategy derived from MPDQN outperforms that of the genetic algorithm, resulting in lower average AoI for vehicles using MPDQN. This is because MPDQN considers the impact of action allocations for each time slot on subsequent AoI, whereas the genetic algorithm does~not.

\begin{figure}[H]
			\includegraphics[scale=0.5]{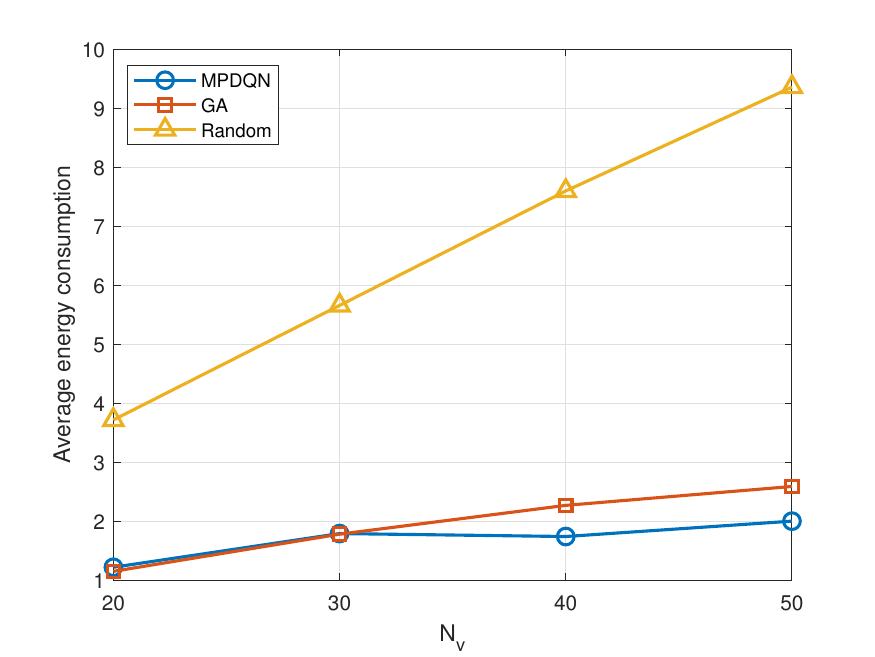}
			\caption{Average energy consumption with $N_v$ under different~algorithms.}
			\label{E}
		\end{figure}

	 Figure~\ref{E} compares the energy consumption within the system when vehicles employ three different methods. It can be observed that energy consumption increases {when the number of devices increases}
		. Moreover, the~energy consumption in the random method does not exhibit significant variations {for different }
		numbers of vehicles within the system, resulting in a more linear energy consumption pattern. On~the other hand, in~MPDQN and genetic algorithm methods, the~increase increase in
		the number of vehicles leads to an increase in 
		interference power, resulting in a decrease in SINR. The~reduced SINR leads to a longer transmission time and a higher {probability} 
		of exceeding transmission time slots, thereby increasing the average information age of the system. However, due to the relatively larger weight of average energy consumption in the optimization objective, RSUs may choose to incur minimal additional energy consumption when the information age is low. Thus, the~impact of the number of vehicles on energy consumption is less pronounced compared to its impact on information age.
		Furthermore, the~strategies obtained by MPDQN and the genetic algorithm consistently outperform the random strategy, as~they can derive better actions to ensure lower energy consumption costs at low AoI through their respective optimization approaches. Additionally, it can be observed that MPDQN consistently outperforms the genetic algorithm when the number of vehicles is high. This is because MPDQN has an advantage in handling real-time decision-making and dynamic environments, allowing it to continuously adjust strategies based on environmental feedback, resulting in stronger adaptability. In~scenarios with a higher number of vehicles, MPDQN may have an easier time learning optimal scheduling strategies through interaction with the~environment.

\begin{figure}[H]
			\includegraphics[scale=0.5]{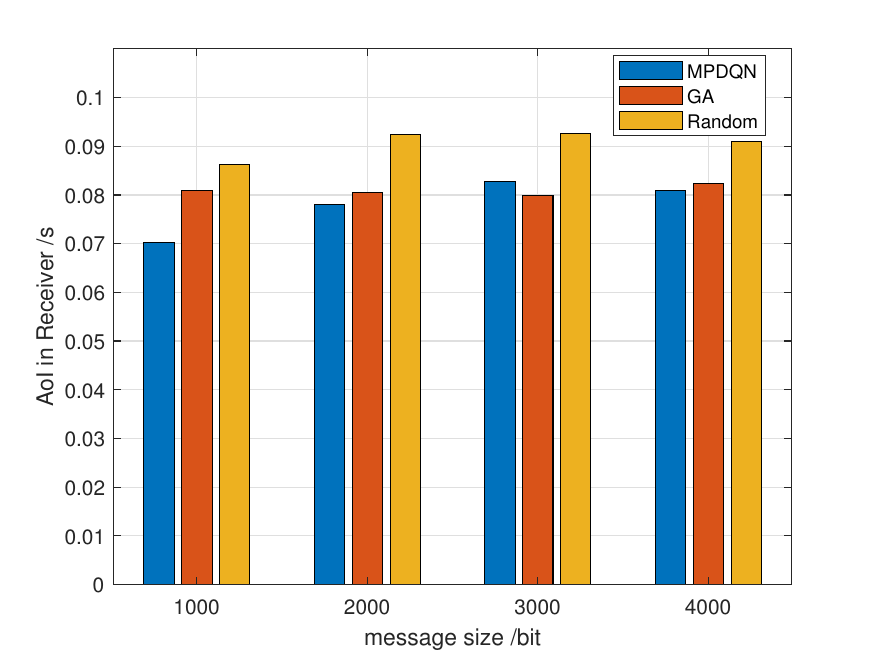}
			\caption{Average AoI with message size under different~algorithms.}
			\label{bitA}
		\end{figure}

		{Figure \ref{bitA} compares the impact of different algorithms on average AoI in a scenario with 50 vehicles. It can be observed that as message size increases, average AoI also tends to increase. This is because the larger size of messages requires higher transmission rates, necessitating greater bandwidth and higher SINR. Among~the three algorithms depicted above, MPDQN generally achieves the lowest average AoI, followed by the GA algorithm, highlighting the effectiveness of MPDQN.}
		
\begin{figure}[H]
			\includegraphics[scale=0.5]{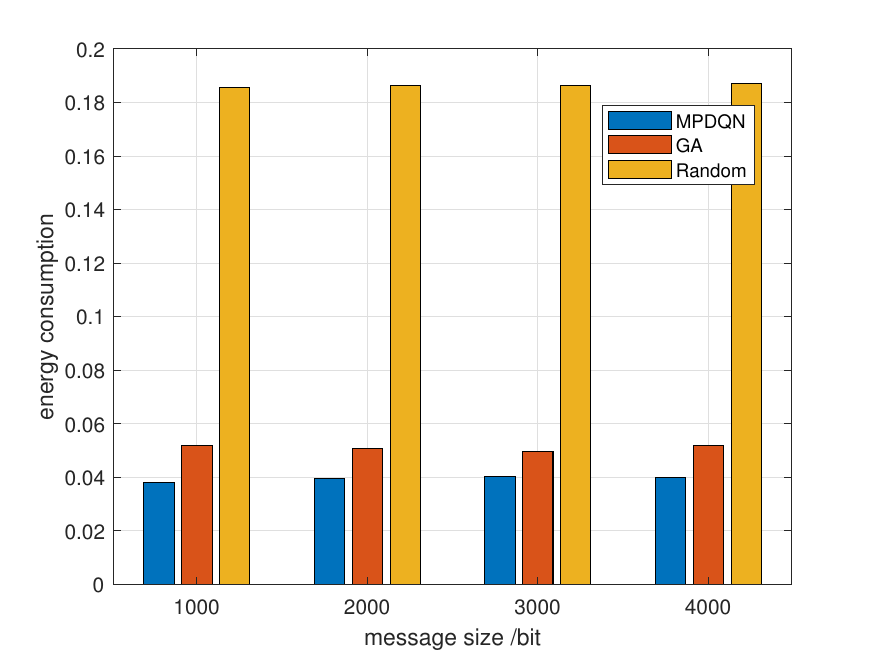}
			\caption{Average energy consumption with message size under different~algorithms.}
			\label{bitE}
		\end{figure}

			{Figure \ref{bitE} compares the influence of different algorithms on average energy consumption in the same scenario as Figure~\ref{bitA}. Here, energy consumption is averaged over the number of vehicles depicted in Figure~\ref{E}. Consistent with the previous findings, MPDQN outperforms GA and random algorithms, ensuring lower energy consumption during vehicle communication in the scenario.}
			
\unskip

		\section{Conclusions}\label{sec6}
		This paper addresses the {probability} 
		of resource collisions in NR-V2X Mode 2 communication, despite the use of autonomous resource selection and probability reselection mechanisms. To~mitigate the impact of collisions on the communication process, we proposed utilizing NOMA's serial interference cancellation mechanism. Additionally, we employed the MPDQN algorithm to dynamically adjust the transmission interval and transmission power of vehicles to reduce the average information age and energy consumption in the system. Firstly, we established communication models for NR-V2X and NOMA and~then constructed a reinforcement learning framework based on MPDQN. In~{the learning framework, }
		we modified the action space to enable simultaneous scheduling of discrete and continuous actions and~finally optimized the joint problem of information age and energy consumption. Through simulation analysis, we demonstrated the advantages of NR-V2X over LTE-V2X, the~improvement effect of NOMA on the information age performance in NR-V2X scenarios, and~the effectiveness of MPDQN in reducing the information age and energy consumption in NR-V2X scenarios. 
		{Some potential challenges in this direction are noted: Future vehicles may use both LTE-V2X and NR-V2X for communication, so considering the coexistence  and integration of these two technologies is a challenge~\cite{naik2019ieee}. In~addition, fairness is often considered a key factor in NOMA-related scenarios~\cite{muhammed2019energy}. Therefore, our future research will focus on performance optimization in scenarios where LTE-V2X and NR-V2X coexist and fairness in resource allocation in relevant scenarios with NOMA.
		In addition, MPDQN is a combination of DQN and DDPG, so the optimal performance can be improved by improving them intuitively.
		}
		\vspace{+6pt}
		
		
		\authorcontributions {Conceptualization, S.S., Z.Z. and Q.W.; Methodology, S.S., Z.Z. and Q.W.; Software, S.S. and Z.Z.; Writing---Original Draft Preparation, Z.Z.; Writing---Review and Editing, P.F. and Q.F. All authors have read and agreed to the published version of the manuscript.}
	
	    \funding{This work was supported in part by the National Natural Science Foundation of China under Grant No. 61701197, in~part by the open research fund of State Key Laboratory of Integrated Services Networks under Grant No. ISN23-11, in~part by the National Key Research and Development Program of China under Grant No. 2021YFA1000500(4), and in~part by the 111 project under Grant No.~B23008.}

	\institutionalreview{\hl{ }}

\informedconsent{\hl{ }}

			\dataavailability{Data are contained within the article.} 
	
	
			\conflictsofinterest{Author Qiang Fan was employed by the company Qualcomm. The~remaining authors declare that the research was conducted in the absence of any commercial or financial relationships that could be construed as a potential conflict of~interest.}

	\begin{adjustwidth}{-\extralength}{0cm}

\reftitle{References}

		\PublishersNote{}
\end{adjustwidth}
	\end{document}